\def\year{2019}\relax
\documentclass[letterpaper]{article} 
\usepackage{aaai19}  
\usepackage{times}  
\usepackage{helvet}  
\usepackage{courier}  
\usepackage{url}  
\usepackage{graphicx}  
\usepackage{amsmath}
\usepackage{tipa}
\usepackage{multirow}
\usepackage{comment}
\usepackage{booktabs}
\usepackage{xcolor}

\newcommand{\citet}[1]
{\citeauthor{#1}~\shortcite{#1}}
\newcommand{\citep}{\cite}

\DeclareMathOperator*{\argmax}{argmax}

\usepackage{url}
\begin{document}
%

\newcommand{\todo}[1]{\textbf{\textcolor{red}{TODO: #1}}}
\newcommand{\cmt}[1]{\textbf{\textcolor{blue}{Comment: #1}}}
\newcommand{\gn}[1]{\textbf{\textcolor{cyan}{[GN: #1]}}}
\newcommand{\sr}[1]{\textbf{\textcolor{orange}{[SR: #1]}}}
\newcommand{\jc}[1]{\textbf{\textcolor{purple}{[JC: #1]}}}

\title{Zero-shot Neural Transfer for Cross-lingual Entity Linking}

\author{Shruti Rijhwani \qquad Jiateng Xie \qquad Graham Neubig \qquad Jaime Carbonell \\
Language Technologies Institute\\
  Carnegie Mellon University \\
  {\tt \{srijhwan, jiatengx, gneubig, jgc\}@cs.cmu.edu} \\}

\date{}
\maketitle
\begin{abstract}

Cross-lingual entity linking maps an entity mention in a source language to its corresponding entry in a structured knowledge base that is in a different (target) language. While previous work relies heavily on bilingual lexical resources to bridge the gap between the source and the target languages, these resources are scarce or unavailable for many low-resource languages. 
To address this problem, we investigate \emph{zero-shot} cross-lingual entity linking, in which we assume no bilingual lexical resources are available in the source low-resource language. 
Specifically, we propose \emph{pivot-based entity linking}, which leverages information from a high-resource ``pivot" language to train character-level neural entity linking models that are transferred to the source low-resource language in a zero-shot manner. With experiments on 9 low-resource languages and transfer through a total of 54 languages, we show that our proposed pivot-based framework improves entity linking accuracy 17\% (absolute) on average over the baseline systems, for the zero-shot scenario.\footnote{All data, resources and code will be made available as a new benchmark for zero-shot cross-lingual entity linking.} Further, we also investigate the use of language-universal \emph{phonological representations} which improves average accuracy (absolute) by 36\% when transferring between languages that use different scripts.

\end{abstract}

\section{Introduction}
\label{sec:intro}

\begin{figure}
    \centering
    \includegraphics[width=1.0\columnwidth]{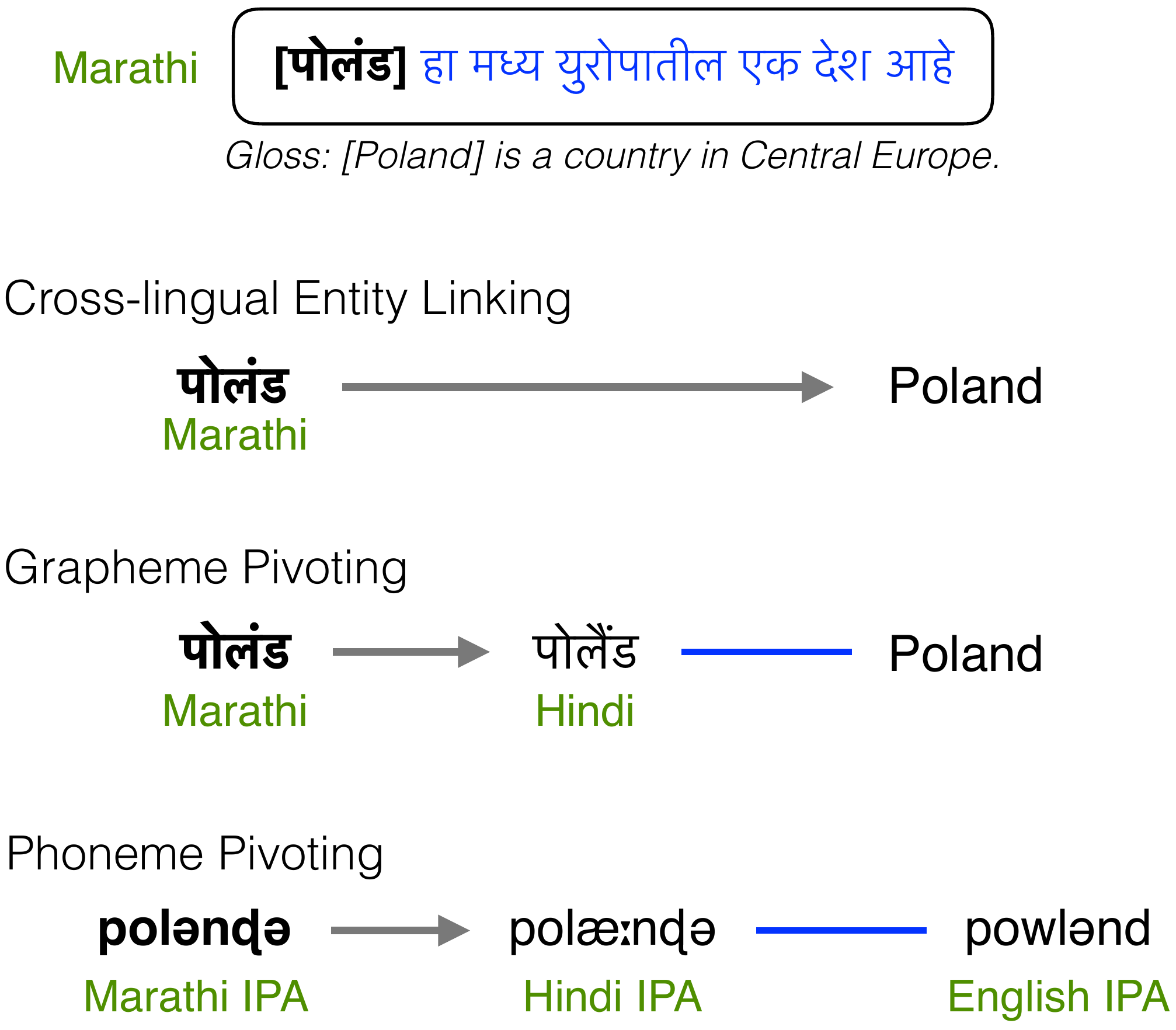}
    \caption{Cross-lingual EL of an entity mention in a Marathi sentence to an English KB: (1) direct Marathi--English linking. (2) Pivoting Marathi through Hindi for linking. (3) Using IPA for pivoting. The arrows represent entity linking and the solid blue lines represent parallel data from bilingual lexicons, used for pivoting.}
    \label{fig:example}
\end{figure}

Entity linking (EL) is the task of associating an entity mention with its corresponding entry in a structured knowledge base (such as Wikipedia or Freebase), with several downstream applications including document understanding, entity and event coreference, text mining and information retrieval~\cite{Mihalcea:2007:WLD:1321440.1321475,han-sun:2012:EMNLP-CoNLL}.
In this work, we focus on cross-lingual EL~\cite{mcnamee-EtAl:2011:IJCNLP-2011}, where the given entity mention is in a (source) language different from the (target) language of the knowledge base. In Figure~\ref{fig:example}, for example, the input entity in Marathi (`Poland') is linked to the appropriate entry in an English knowledge base (KB).

In monolingual EL, simple methods like string similarity and Wikipedia anchor-text can be used effectively to identify candidate KB entries for each entity mention~\cite{ji-grishman:2011:ACL-HLT2011,sil2017neural}. However, such methods often fail in the case of cross-lingual EL, because entity mentions in the source language and KB entries in the target language are frequently dissimilar. Existing work uses bilingual resources to bridge this gap, including lexicons and Wikipedia inter-language links~\cite{tsai-roth:2016:N16-1,pan-EtAl:2017:Long2,TsaiRo18}. However, the vast majority of the world's $\approx 7,000$ living languages are low-resourced, and have extremely limited or zero bilingual resources. Even within the 300 languages available on Wikipedia, some have an extremely small number of articles (for example, Oromo and Tigrinya have only 773 and 168 Wikipedia pages respectively, while English has over 5 million). In order to enable cross-lingual EL for such low-resource languages (LRLs), it is imperative to design methods that do not rely heavily on lexicons or other resources in the LRLs. 

In this work, we take this to the extreme: we perform the first study on \emph{zero-shot} cross-lingual entity linking, devising methods that require \emph{no} bilingual resources in the source language (i.e., the language that the input entity mention is sourced from). We propose pivot-based entity linking, or \textsc{Pbel}, which is based on the intuition that despite the fact that many languages have very few resources, it is common that these languages have closely-related higher-resource languages (HRLs) that we can leverage. For example, the relatively low-resourced languages of Marathi and Lao are from the same language family as the high-resourced Hindi and Thai respectively. We exploit bilingual lexicons and structured information available in these HRLs to improve EL for languages where no such resources are available. The specific contributions of this paper are

\begin{itemize}
    \item \textsc{Pbel}, a novel method for cross-lingual EL, that uses \emph{no} bilingual resources in the source language. This consists of two components:
    
    \textbf{Zero-shot transfer of neural entity linking models:} Using a bilingual lexicon between an HRL and English, we train a character-level neural model for linking entities in the HRL to an English KB. The model can be transferred to perform EL for a source LRL, without any language-specific fine-tuning. For example, we train a model to link Hindi (HRL) to English and transfer the model to link Marathi (LRL) to English. Such transfer learning schemes have been successful for other tasks, such as morphological tagging and machine translation, when used between closely-related languages~\cite{zoph-EtAl:2016:EMNLP2016,cotterell-heigold:2017:EMNLP2017}.

    \textbf{Pivoting:} 
    Rather than attempting to directly link an LRL entity to English, we link the entity to a closely-related HRL. We then use bilingual lexicons, readily available in the HRL, to obtain the corresponding English entity link. We are, therefore, using the HRL as an intermediate \emph{pivot} between the source LRL and English. Our experiments demonstrate that this significantly improves EL accuracy over directly linking to English, as named entities are likely similar in related languages~\cite{Tsvetkov2016CrossLingualBW}. In Figure~\ref{fig:example}, the orthographic pivoting example shows that `Poland' in Marathi and Hindi are written similarly and can be linked with our neural EL model. Since we have extensive Hindi-English lexicons, we can obtain the English KB entry quite simply from the Hindi name (shown with a solid blue line in the figure).
    
    \item The use of \emph{phonological representations} for cross-lingual EL. Transfer of character-level models is bound to fail when the HRL and LRL do not use the same writing system. We propose using International Phonetic Alphabet (IPA) to bridge the gap between different scripts. Figure~\ref{fig:example} shows an example of pivoting in the phonological space. We see that the pronunciation (IPA representation) of `Poland' is highly similar in Marathi and Hindi because they are closely-related languages.
    
    \item Experiments with 9 test languages from various language families and 54 transfer languages that analyze the performance of current state-of-the-art cross-lingual entity linking methods in truly zero-shot settings, in order to demonstrate the effectiveness of the proposed \textsc{Pbel} method.
    
\end{itemize}


\section{Problem Setting}
\label{sec:el}

Cross-lingual EL is the task of linking an entity mention $m$ in a source language to a structured KB $\mathcal{K}$ in a target language. In our work, we study the case when the source language is some low-resource language (LRL), and follow most previous work by using an English KB as the target~\cite{pan-EtAl:2017:Long2,TsaiRo18}. The task involves identifying the appropriate entry $e_\texttt{en} \in \mathcal{K}$ that corresponds to the entity mention $m$. Our EL model predicts an entity link by maximizing a score function between $m$ and $e_\texttt{en}$.
$$\hat{e}_\texttt{en} = \argmax_{e_\texttt{en} \in \mathcal{K}} \text{score}(m, e_\texttt{en})$$

In addition, consider an HRL that is closely-related (via language family, script, similar phonology etc.) to the source LRL, for which bilingual resources with links to the English KB are available. These are easily obtainable from inter-language links in massive multilingual resources like Wikipedia, DBPedia~\cite{Auer:2007:DNW:1785162.1785216} and BabelNet~\cite{NavigliPonzetto:12aij}. For each entry $e_\texttt{en} \in \mathcal{K}$, let the parallel entity in the HRL be $e_\texttt{HRL}$. Note that it is possible that $e_\texttt{HRL} = \emptyset$ for some $e_\texttt{en}$, as not all English entities are language-linked to the HRL. These HRL parallel entities are the only cross-lingual resources we have access to in this work. They are used for training the EL model as well as for pivoting, as described in the following section. We do not use \emph{any} parallel entities in the LRL.

\section{Model}
\label{sec:model}

The basic component of our entity linking system consists of two neural encoders, one for the HRL and one for English, which convert named entities (character sequences) into vector representations. The two encoders are trained such that the vector representations of two parallel entities $e_\texttt{HRL}$ and $e_\texttt{en}$ are similar. We use the two encoders to calculate two sets of scores between an LRL entity mention $m$ and a KB entry $e_\texttt{en}$ through the following methods, leveraging the fact that the LRL and HRL are closely-related:

\begin{itemize}
    \item \textbf{Zero-shot Transfer} We apply the HRL encoder directly on $m$ and the English encoder on $e_\texttt{en}$, and calculate a score based on the similarity of the vector representations.
    \item \textbf{Pivoting} We apply the HRL encoder on both the LRL entity mention $m$ and the parallel entry $e_\texttt{HRL}$, and calculate a score based on the similarity of the vector representations.
\end{itemize}
Then, we take the maximum out of these two scores as the score between $m$ and $e_\texttt{en}$.

In the following sections, we describe each part of the model in more detail, and discuss using phonological representations. For simplicity, we introduce the system with a single HRL, and discuss transfer from multiple HRLs later.

We should note that, in most existing work, the EL process is a two-step pipeline -- candidate retrieval and context-based disambiguation~\cite{Hachey:2013:EEL:2405838.2405914}. Candidate retrieval reduces the search space for EL by selecting a small number of candidates for more precise linking. This is required because precise linking algorithms are often prohibitively expensive to use on the entire KB. Our proposed method is context-insensitive, much like traditional candidate retrieval models. Although we present our method as an end-to-end EL system by simply picking the top-scoring KB entry as the entity link, we also discuss how it can be used to retrieve a larger number of candidates in next section. 

\subsection{Entity Similarity Encoder}
\label{sec:encoder}

\begin{figure}
    \centering
    \includegraphics[width=0.9\columnwidth]{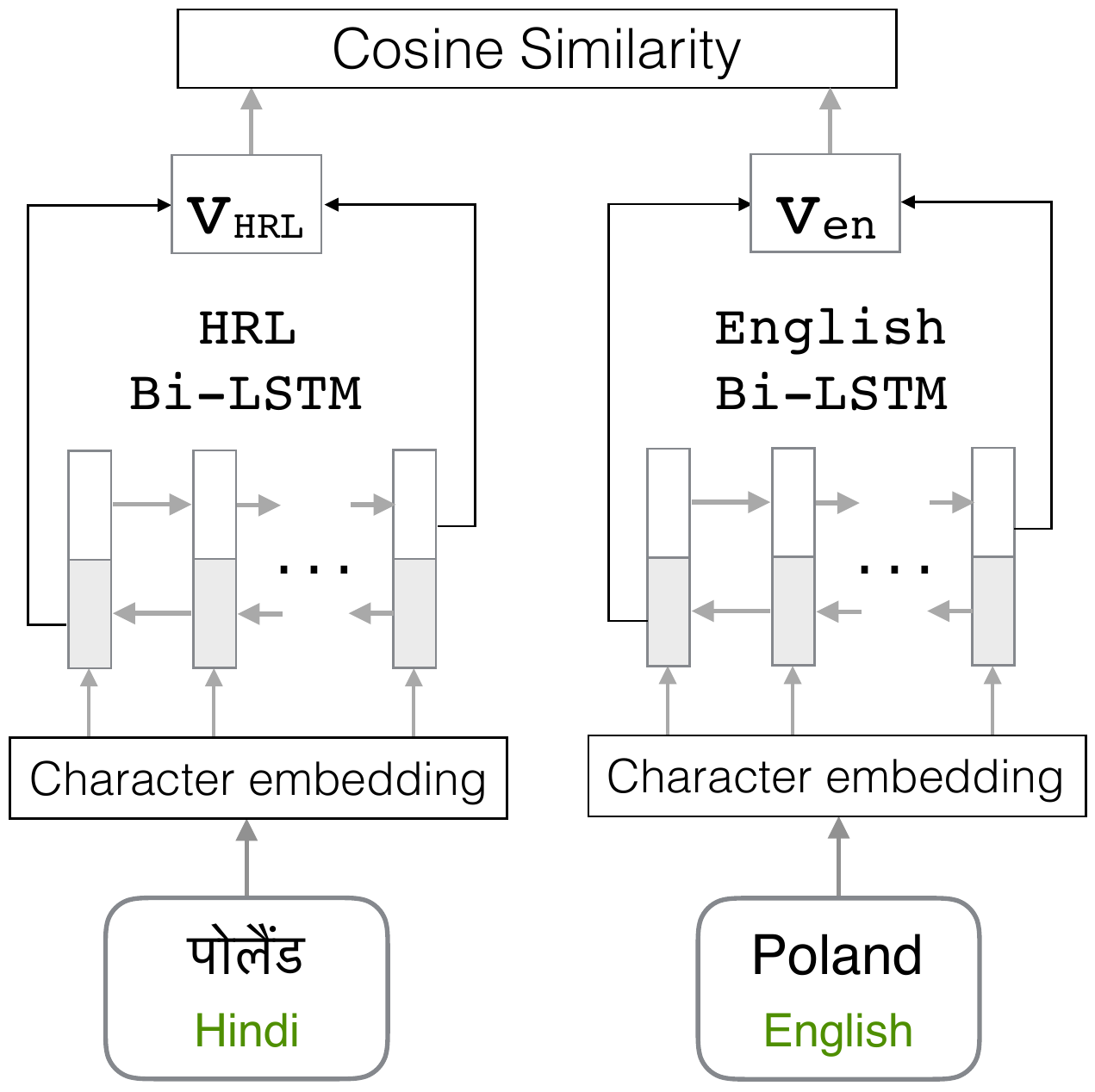}
    \caption{Entity similarity model trained on parallel entities between an HRL and English. An example entity (`Poland') with Hindi as the HRL is shown here.}
    \label{fig:encoder}
\end{figure}

The entity similarity model is shown in Figure~\ref{fig:encoder}. We use character-level Bidirectional-LSTMs (Bi-LSTM) to encode entities into a continuous vector space. The model is trained to maximize the cosine similarity between the vector representations of an entity in an HRL and its equivalent (i.e., parallel) entity in English, much like recent work in neural information retrieval~\cite{mitra2017neural}.

Consider an entity $e_\texttt{HRL}$ and its parallel entity in English $e_\texttt{en}$. Each entity is a sequence of characters; $e_\texttt{HRL} = \langle c_1, c_2 \dots c_M \rangle $ and $e_\texttt{en} = \langle k_1, k_2 \dots k_N \rangle $. For each character, we obtain a fixed-size character embedding. 
The embeddings are used as input to the Bi-LSTM and the final states (concatenation of the last states from the forward and backward LSTMs) form the encoded entity vectors $\mathbf{v_\texttt{HRL}}$ and $\mathbf{v_{\texttt{en}}}$. 
$$\mathbf{v_\texttt{HRL}} = \texttt{HRL-Bi-LSTM}(\langle c_1, c_2 \dots c_M \rangle)$$
$$\mathbf{v_{\texttt{en}}} = \texttt{English-Bi-LSTM}(\langle k_1, k_2 \dots k_N \rangle)$$
The similarity score is computed as, 
$$\text{sim}(e_\texttt{HRL}, e_{\texttt{en}}) = \text{cosine}(\mathbf{v_\texttt{HRL}}, \mathbf{v_{\texttt{en}}})$$

Since we want to efficiently train a model that can rank KB entries for a given mention, we follow existing work and use negative sampling with a max-margin loss for training the encoder~\cite{collobert2011natural}. The loss function is 
$$L = \text{max}(0, \text{sim}(e_\texttt{HRL}, e_{\texttt{en}})-\text{sim}(e_\texttt{HRL}, e^*_{\texttt{en}})+\lambda)$$
where $\lambda$ is the margin and $e^*_{\texttt{en}}$ is a negative KB example such that $e \neq e^*$.\footnote{For each mini-batch of training data consisting of paired HRL-English entities, we use the correct KB entry as the positive sample, and all other KB entries in the mini-batch as negative samples.} 
\medskip

\subsection{Zero-shot Transfer to LRL}

If we train the model using an HRL that is sufficiently similar to the source LRL, the entity encoder can effectively predict similarity between an entity mention $m$ and the English KB entries. We use the learned \texttt{HRL-Bi-LSTM} to encode $m$ and the \texttt{English-Bi-LSTM} to encode an English KB entry $e _{\texttt{en}}\in \mathcal{K}$. The score for $e_{\texttt{en}}$ is the cosine similarity between these encodings:
$$\text{sim}(m, e_{\texttt{en}}) = \text{cosine}(\mathbf{v_\texttt{m}}, \mathbf{v_\texttt{en}})$$
where $\mathbf{v_\texttt{m}} = \texttt{HRL-Bi-LSTM}(m)$, and $m$ is a sequence of characters.

\subsection{Pivoting}
\label{sec:pivot}

By transferring the entity encoder (as discussed above), we can compute similarity between an LRL mention and the English KB entries. In addition, we propose pivoting, which uses an HRL as an intermediate \emph{pivot} between the LRL and English.
Specifically, instead of considering the English entity $e_\texttt{en}$, we consider its parallel entity $e_\texttt{HRL}$ in the HRL. We use the learned \texttt{HRL-Bi-LSTM} to encode both $m$ and $e_\texttt{HRL}$, and calculate a similarity score between the encodings:
$$\text{sim}(m, e_\texttt{HRL}) = \text{cosine}(\mathbf{v_\texttt{m}}, \mathbf{v_\texttt{HRL}})$$
Note that this score is used for $e_\texttt{en}$, as $e_\texttt{en}$ and $e_\texttt{HRL}$ refer to the same entity (in English and the pivot HRL respectively).

\smallskip




\begin{figure}
    \centering
    \includegraphics[width=\columnwidth]{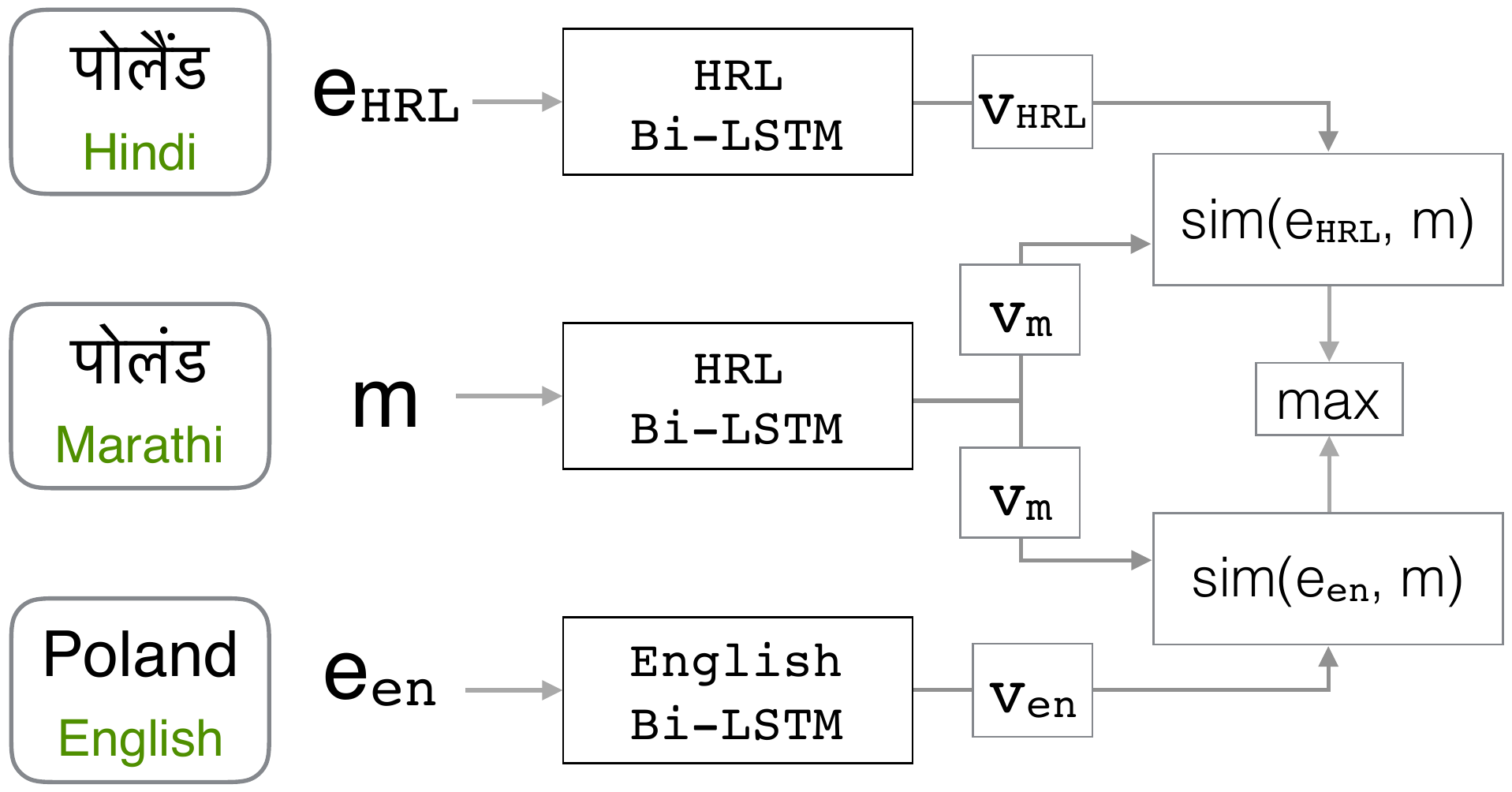}
    \caption{Architecture to compute $\text{score}(m, e_{\texttt{en}})$ with pivoting through a high-resource language entity $e_\texttt{HRL}$. An example entity (`Poland') is shown here, with Marathi as the source LRL and Hindi as the pivot HRL.}
    \label{fig:pivot}
\end{figure}

\subsection{Calculate EL score}

Using the two methods described above, we have two sets of scores between $m$ and $e_\texttt{en}$, and we take the maximum out of the two as our final score (Figure~\ref{fig:pivot}):
\begin{equation}
    \text{score}(m, e_{\texttt{en}}) = \text{max}(\text{sim}(m, e_\texttt{en}), \text{sim}(m, e_\texttt{HRL}))
    \label{eq: pivot}
\end{equation}

The score function for EL is modified to consider both the similarity between $m$ and $e_\texttt{en}$ and the similarity between $m$ and $e_\texttt{HRL}$. The final score for $e_{\texttt{en}} \in \mathcal{K}$ takes the maximum of the two scores because our objective is to maximize the similarity. Since not all the entries in $\mathcal{K}$ have a parallel entity in \texttt{HRL}, it is non-trivial to use another combination of the scores (instead of max). If there is no parallel entity, we consider $\text{sim}(m, e_\texttt{HRL}) = -\infty$.

\subsection{Phonological Representation}
\label{sec:phon}

Since the entity similarity model is a character-level neural network, cross-lingual transfer is bound to fail if the LRL uses a different script than the HRL used for training the encoder. To overcome this problem, we propose training the encoder in the language-universal phonological space. We experiment with two representations:

\smallskip
\noindent
\textbf{Phoneme embeddings}: We convert all parallel training data between the HRL and English into IPA using Epitran, a grapheme-to-phoneme system that supports over 55 languages~\cite{MORTENSEN18.890}. The encoder model itself is agnostic to the characters used to represent the entities and can be trained with the IPA parallel data in the same manner as described above.
\smallskip

\noindent 
\textbf{Articulatory feature embeddings}:
We transform our parallel IPA training data into articulatory feature sets using PanPhon~\cite{panphon}. Articulatory features can potentially capture important characteristics of the pronunciation that may not be apparent from the IPA, as indicated by improved low-resource Named Entity Recognition by~\cite{panphon}.
PanPhon converts each IPA segment into 21 features, which represent phonological aspects of the input (including voice, nasal, strident etc.) Since each word is a sequence of IPA segments, we obtain a sequence of feature vectors from PanPhon. These features are linearly transformed into an embedding (which replaces the character embedding in Figure~\ref{fig:encoder}) using a weight matrix and bias vector, both of which are trained with the encoder model.

\subsection{Phylogenetic Weighting}
\label{sec:multi}

So far, we have discussed transferring to the LRL from a single HRL. However, we can easily train multiple encoders on different HRLs. In order to leverage these models, we use a simple weighting mechanism, whereby the similarity scores from each HRL model is weighted by the \textit{phylogenetic distance} between the HRL and the source LRL. The KB entry with the maximum score after weighting is selected as the predicted entity link. Phylogenetic distance, obtained from the URIEL Database~\cite{littell2017uriel}, represents the relative distance between two languages in the Glottolog hypothesized tree of language~\cite{glottolog}. This essentially implies languages that have similar origin will be closer in distance, which is intuitively useful for identifying appropriate \textit{pivot} HRLs for a specific LRL.

\section{Experiments}
\label{sec:experiment}

\begin{table*}[tb]
\centering
\renewcommand{\arraystretch}{1.07}
\resizebox{\textwidth}{!}{
\begin{tabular}{@{}l|rrrrrrrrr|c@{}}
  Model & \multicolumn{1}{c}{bn} & \multicolumn{1}{c}{jv} & \multicolumn{1}{c}{lo} & \multicolumn{1}{c}{mr} & \multicolumn{1}{c}{pa} & \multicolumn{1}{c}{te} & \multicolumn{1}{c}{ti} & \multicolumn{1}{c}{uk} & \multicolumn{1}{c|}{ug} & Avg.\\ \midrule
  \textbf{\textsc{Exact}} & 
   .00 & .63 & .02 & .00 & .00 & .00 & .02 & .02 & .03 & .08\\
  
  \textbf{\textsc{Trans}} & 
   .00 & .63 & .02 & .17 & .00 & .00 & .46 & .02 & .03 & .15\\ \midrule
  \textbf{\textsc{Encode}}  & &  &&&&&&&  \\
  \textsc{Manual}
  & .36 (hi) & .70 (id) & .07 (th) & .46 (hi) & .31 (hi) & .20 (ta) & .44 (am) & .25 (ru) & .16 (tr) & .33\\
   \textsc{Best-53}
   & .38 (ms) & .70 (id) & .07 (th) & .46 (hi) & .36 (te) & .36 (pa) & .44 (am) & .41 (kk) & .16 (tr) & .37\\
  \midrule
    \textbf{\textsc{Pbel}} & &  &&&&&&&  \\
  \textsc{Manual}
 & .48 (hi) & .86 (id) & \textbf{.28} (th) & \textbf{.62} (hi) & \textbf{.49} (hi) & .33 (ta) & \textbf{.69} (am) & .50 (ru) & .32 (tr) & .51 \\
   \textsc{Best-53}
 & .48 (hi) & .86 (id) & \textbf{.28} (th) & \textbf{.62} (hi) & \textbf{.49} (hi) & \textbf{.47} (hi) & \textbf{.69} (am) & .54 (kk) & .32 (tr) & .53\\
  \textsc{Multi} & 
   \textbf{.53} & \textbf{.87} & \textbf{.28} & \textbf{.62} & .48 & .46 & \textbf{.69} & \textbf{.56} & \textbf{.40} & \textbf{.54} \\ 
\end{tabular}
}
\caption{Accuracy for cross-lingual Wikipedia title linking, with the transfer HRL shown in parentheses. The best accuracy among input representations with graphemes, phonemes or articulatory features for \textsc{Encode} and \textsc{PBEL} is presented here. Complete results for each representation are in the supplementary material. }
\label{table: wiki}
\end{table*}


In this section, we discuss our experimental setting, baselines and results on several low-resource languages. We attempt to answer the following research questions: (1) ``Does the proposed \textsc{Pbel} method outperform methods that do not perform pivoting?'' (2) ``What is the interplay of the orthographic or phonological input representation with the features of the high-resource transfer language and the low-resource test language?'' (3) ``What is the potential of using pivoting for candidate retrieval in 2-step EL systems?'' (4) ``How much does a small amount of lexical data in the LRL improve \textsc{Pbel}?''

\subsection{Experimental Settings}

In order to comprehensively, yet realistically, examine performance on a wide variety of low-resource language pairs, we perform experimental validation on two varieties of the task.
For both, we measure the performance of the baselines and our system with linking accuracy -- i.e. the fraction of instances when the system-predicted link was the ``true'' KB entry for the given input.

\subsubsection{Cross-lingual KB Title Linking}

Our test set is constructed from Wikipedia parallel titles between the LRL and English. That is, the `gold-standard' link for an article title in the LRL is the corresponding English Wikipedia entry. Note that these parallel titles are used \emph{only for testing}.

In contrast to the traditional EL task of linking textual entity mentions to KB entries, this experimental setting is similar to ``Cross-lingual Article Linking''~\cite{wang-wu-tsai2014}, where we link Wikipedia article titles in the LRL to English KB entries. We focus on articles about named entities (locations, persons and organizations). Most work~\cite{sorg2008enriching,wang2012cross,wang-wu-tsai2014} in cross-lingual article linking assumes the availability of a KB in the source LRL. This is unrealistic for our zero-shot scenario and hence, we compare with existing methods that do not rely on this assumption.

We test on nine relatively low-resource languages from various language families: Tigrinya (ti), Lao (lo), Uyghur (ug), Telugu (te), Punjabi (pa), Javanese (jv), Marathi (mr), Bengali (bn) and Ukrainian (uk). We have 2000 titles in the test set for each language, apart from ti, lo and ug, for which we have 90, 579 and 1297 instances respectively\footnote{Due to the small size of Wikipedia in these languages.}. The target KB is English Wikipedia, which contains 2.1 million titles. We use a total of 54 HRLs as potential transfer languages, details of which are in the supplementary material.

We primarily use Wikipedia links as a test bed because of the availability of data in many LRLs, not found in traditional EL datasets. We note that although the title linking task is not identical to entity mention linking, it maintains much of the difficulty associated with cross-lingual EL, particularly with respect to how state-of-the-art translation-based techniques perform poorly in the zero-resource setting. However, the challenge of linking ambiguous mentions is eliminated in this task, which is why we also test our method on a full cross-lingual EL dataset, described below. 

\subsubsection{Full Cross-lingual Entity Linking}
We also test our proposed \textsc{PBEL} method on the standard cross-lingual EL setting of linking textual mentions to KB entries. For the test set, we use annotated documents from the DARPA LORELEI program\footnote{\url{https://www.darpa.mil/program/low-resource-languages-for-emergent-incidents}}, in two extremely low-resource languages -- Tigrinya and Oromo. These are news articles, blogs and social media posts about disasters and humanitarian crises (floods, famine, political unrest etc.). The data contains named entity mentions extracted from the texts, annotated with their respective KB links. The English KB provided with the dataset contains over 11M entries. This KB is much larger than English Wikipedia and includes entities from, among other sources, GeoNames and the CIA World Factbook. 



\subsection{Entity Similarity Scoring Models}

We consider three models for scoring KB entries for cross-lingual EL.
Two are based on existing literature on state-of-the-art monolingual or cross-lingual EL methods~\cite{ji-grishman:2011:ACL-HLT2011,pan-EtAl:2017:Long2,sil2017neural}, which have been shown to work in the \emph{supervised} setting where an entity lexicon for the LRL is available, but are intuitively less suited for our zero-shot setting.
The third is the character-level neural decoder described in the previous section, which we posit is more appropriate for zero-shot transfer, and for use with our proposed \emph{pivoting} method.

\begin{itemize}

\item \textsc{Exact}: Exact match to the KB is used in state-of-the-art monolingual EL systems~\cite{sil2017neural}.\footnote{Monolingual EL also often uses Wikipedia anchor text for matching, which is infeasible in the LRL scenario.} The predicted entity link, if found, is the KB entry that is an exact string match with the mention $m$.
\smallskip

\item \textsc{Trans}: This baseline is a candidate retrieval technique used in the state-of-the-art low-resource EL system by~\citet{pan-EtAl:2017:Long2}, which attempts to translate the mention $m$ into English in order to predict the entity link. We generate a bilingual lexicon with word alignments on parallel Wikipedia titles using \texttt{fast\_align}~\cite{dyer-chahuneau-smith:2013:NAACL-HLT}. Each word in the input entity $m$ is translated to English using the lexicon. The predicted entity link is the exact match of the obtained translation, if found, in $\mathcal{K}$. We experiment with two varieties of lexicon creation with different resource requirements:

\textbf{Supervised.}\quad The lexicon is created from a small number of parallel entities between the LRL and English.

\textbf{Zero-shot.}\quad The lexicon is created from parallel entities between a closely-related HRL and English.

\item \textsc{Encode}: We train a similarity encoder, as seen in Figure~\ref{fig:encoder}, using parallel Wikipedia titles between English and an HRL.
We transfer the trained \texttt{HRL-Bi-LSTM}, without fine-tuning, to encode the mention $m$. The \textsc{Encode} model does not make use of ``pivoting" and directly compare $m$ with the English KB entries. 
These models are trained in either orthographic or phonological space.

The entity similarity encoder model is implemented in DyNet~\cite{dynet}, with a character embedding size of $64$ and LSTM hidden layer size of $512$.

\end{itemize}

\subsection{Results: Cross-lingual KB Title Linking}

As described above, we test on nine LRLs. We experiment with variants of the models that differ in terms of selecting an appropriate HRL for transfer:

\begin{itemize}
    \item \textsc{Manual}: We manually choose an HRL a priori, which is closely related (from the same language family) to the source LRL. The HRLs we select are (for the respective LRLs) -- Amharic (Tigrinya), Thai (Lao), Turkish (Uyghur), Tamil (Telugu), Hindi (Punjabi, Marathi, Bengali), Indonesian (Javenese) and Russian (Ukrainian). The HRLs selected are of different script than the LRL for some languages (Lao-Thai, Bengali-Hindi etc.), in order to test the utility of phonological transfer.
    \item \textsc{Best-53}: We attempt to transfer similarity encoders trained on 53 potential HRLs and present the HRL that obtained the highest accuracy for each of the nine test languages. The 53 are all the languages supported by Epitran~\cite{MORTENSEN18.890}, apart from the source LRL itself.
    
    \item \textsc{Multi}: In this setting, we use multiple pivot languages for a single source LRL. We use all 53 languages \emph{together}, and experiment with both unweighted and phylogenetic-distance-weighted combination, as described in the previous section.
\end{itemize}

We use the above methods with the \textsc{Encode} model as well as our proposed \textsc{Pbel} method. For \textsc{Encode}, we use the HRL only for training the encoder. For \textsc{Pbel}, we additionally use the HRL for \emph{pivoting}, i.e., to score KB entries as described in Equation~\ref{eq: pivot}. We also compare with \textsc{Exact} and \textsc{Trans} (Zero-shot). The supervised \textsc{Trans} uses a lexicon built from Wikipedia parallel titles in the LRL, which we also use to construct the test set, leading to an unfair comparison for the cross-lingual title linking task.

The entity linking accuracy on the Wikipedia test set are summarized in Table~\ref{table: wiki}. On average, our proposed \textsc{Pbel} method performs significantly better than the baselines, with significant accuracy gains in all nine test languages.

The \textsc{Exact} baseline, which is most often used for monolingual EL~\cite{sil2017neural}, performs reasonably only when the test language is in the same script as English (i.e., Javanese). Similarly \textsc{Trans}, the current state-of-the-art retrieval method in cross-lingual EL~\cite{pan-EtAl:2017:Long2}, fails when zero data is available in the test language, unless the HRL is very closely-related to the LRL (as with \textit{jv}, \textit{mr} and \textit{am}). On the other hand, \textsc{Encode} presents relatively strong zero-shot transfer results.

\textsc{Pbel} offers stronger performance than \textsc{Encode} because it considers similarity of the LRL mention with both the HRL and English. As seen in the \textsc{Best-53} results, the HRL that performs best is closely-related to the respective test LRL (language family, shared writing system or geographic proximity). Since \textsc{Pbel} leverages this similarity, it becomes easier to predict the correct link. We also observe that using multiple pivot HRLs leads to better average accuracy, with considerable improvement for some languages.

In most cases, the \textsc{Manual} HRL is also the best performing in \textsc{Best-53}. However, we see that the Dravidian Telugu (te) seems to obtain higher accuracy with Indo-Aryan HRLs -- Punjabi (pa) or Hindi (hi). This could be because Tamil (ta) uses a different script and is relatively distant from Telugu in the Dravidian family~\cite{rama2013distance}. We also see that the Ukrainian (uk) test set has better performance with another Cyrillic script language, Kazakh (kk), rather than Russian (ru). We attribute this to the fact that, on Wikipedia, person names in Russian are written \textit{last-name,first-name}, but \textit{first-name,last-name} in Ukrainian, which reduces the success of zero-shot transfer.

\subsection{Results: Full Cross-lingual Entity Linking}

\begin{table}[tb]
    \centering
    \renewcommand{\arraystretch}{1.05}
    \begin{tabular}{@{}ll|rr@{}}
         Lang. & Tigrinya & Oromo \\\midrule
         \textsc{Exact} & 0.00 & 0.01 \\
         \textsc{Trans} Supervised & 0.21 & 0.05\\ 
         \textsc{Trans} Unsupervised & 0.21 & 0.01 \\
         {\textsc{Encode}} 
         & 0.16 & 0.10\\
         {\textsc{Pbel}} 
         & \textbf{0.33} & \textbf{0.11}\\

    \end{tabular}
    \caption{Entity linking accuracy on non-Wikipedia data}
    \label{tab:nonwiki}
\end{table}

We select appropriate HRLs a priori for training \textsc{Encode} and \textsc{Pbel} -- Amharic for Tigrinya and Somali for Oromo. We compare with the \textsc{Exact}, supervised and unsupervised \textsc{Trans} and \textsc{Encode} scoring models. The training data for the encoders is Wikipedia parallel entities between the HRL and English, as in the previous section.

Entity linking accuracies on the LORELEI dataset are shown in Table~\ref{tab:nonwiki}. \textsc{Pbel} has considerably higher accuracy than the other methods. However, we see relatively lower improvement in accuracy with Somali-Oromo than Amharic-Tigrinya. This is because Somali and Oromo, despite both being Afro-Asiatic languages, are from different sub-language-family branches and are not similar enough for strong transfer performance\footnote{We used Somali because it is the closest language to Oromo that is available on both Wikipedia and Epitran.}~\cite{banti1988two}. The availability of a closely-related HRL is essential to the success of \textsc{Pbel}.

Surprisingly, the supervised \textsc{Trans} model, which uses Wikipedia parallel data in the LRL itself as a lexicon, does not perform better than the zero-shot \textsc{Pbel}. We primarily attribute this to the extremely small size of Wikipedia in these languages (90 Tigrinya and 295 Oromo parallel links with English). We also note that \textsc{Trans} relies solely on lexicon string lookup. This is particularly disadvantageous for Oromo, where there are several accepted spelling variants of the same word and the Wikipedia lexicon contains just one of these (for example `Ethiopia' can be written as `Itiyoophiyaa', `Itoophiyaa', `Itoopiyaa', `Toophiyaa' or `Itophiyaa'). The character-level LSTM we use is likely able to normalize across some of these variations~\cite{lample,luong-manning:2016:P16-1}. Overall, our proposed \textsc{Pbel} method counters several of the limitations of using bilingual lexicons, which strongly affects the performance on such extremely low-resource EL datasets.


\subsection{Analysis}

\begin{table*}[tb]
\centering
\renewcommand{\arraystretch}{1.05}
\begin{tabular}{@{}l|rrrrrrrrr@{}}
  Input & *bn (hi) & jv (id) & *lo (th) & mr (hi) & *pa (hi) & *te (ta) & ti (am) & *uk (ru) & *ug (tr) \\ \midrule
Grapheme & .00 & \textbf{.86} & .02 & \textbf{.62} & .00 & .00 & .61 & \textbf{.50} & .08 \\
Phoneme & \textbf{.48} & .84 & .20 & .58 & .18 & .10 & \textbf{.69} & .23 & .21 \\
Articulatory & .45 & .82 & \textbf{.28} & .56 & \textbf{.49} & \textbf{.33} & .63 & .42 & \textbf{.32} \\

\end{tabular}
\caption{Entity linking accuracy with \textsc{PBEL}, using Graphemes, Phonemes or Articulatory features as input. The HRL used for training and pivoting is shown in parentheses in the first row. The pairs with the different scripts are marked with a ``*''.}
\label{table: rep}
\end{table*}

\subsubsection{Character Representation}
\label{sec: analysis_rep}

We look at the use of different character representations: orthographic (graphemes), IPA (phonemes) and articulatory features. The \textsc{Pbel} model results for our test set with each input representation are presented in Table~\ref{table: rep}. The HRLs used are the same as \textsc{Manual}.

We see that using phonological representations (phonemes and articulatory features) offers the ability to map between languages that use different orthographies, explaining the convincing improvement over graphemes for HRL-LRL pairs that are written in different scripts (Table~\ref{table: rep}). With graphemes, the experiments on these languages achieve $\approx 0$ accuracy because the character vocabulary of the HRL encoder simply does not contain the low-resource test language characters. This is the case with Lao-Thai (\textit{lo}-\textit{th}), Telugu-Tamil (\textit{te}-\textit{ta}) and Bengali-Hindi (\textit{bn}-\textit{hi}). In contrast, we observe that the grapheme representation offers strong transfer performance when the LRL and HRL share orthography, notably Javanese-Indonesian (\textit{jv}-\textit{id}), Marathi-Hindi (\textit{mr}-\textit{hi}) and Ukrainian-Russian (\textit{uk}-\textit{ru}).

\subsubsection{Pivoting for Candidate Generation}

 \label{sec:candidate_gen}
\begin{figure}[tb]
    \centering
    \includegraphics[width=.85\columnwidth]{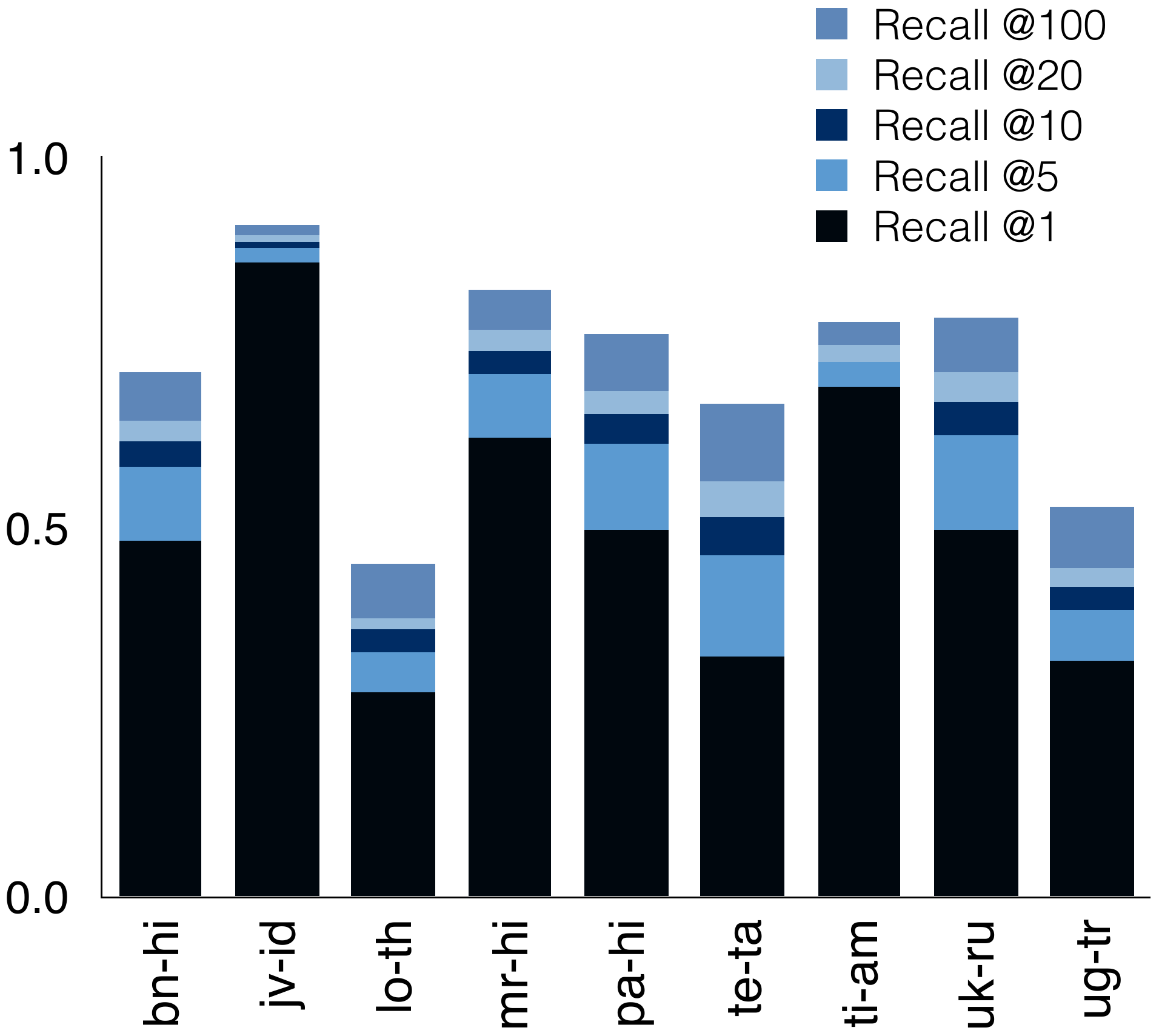}
    \caption{Recall @$k$ for nine test languages}
    \label{fig:candidate}
\end{figure}

 Up to this point, we have demonstrated that our pivoting method can be used for end-to-end EL. However, as mentioned in the previous section, it is common in standard EL systems to use candidate retrieval methods that feed into a more sophisticated downstream disambiguation model. In this case, it is of interest how good the entity linking algorithm is at generating lists of $k$-best candidates. Figure~\ref{fig:candidate} examines this by showing the `Recall @$k$' -- the fraction of instances when the correct entity link is present in the top-$k$ entities as scored by our system~\cite{TsaiRo18}. We observe that for systems with high entity linking accuracy (recall @$1$) like Javanese-Indonesian (\textit{jv}-\textit{id}) and Tigrinya-Amharic (\textit{ti}-\textit{am}), recall at higher $k$ values offer diminishing returns. However, other languages show considerable gain in recall even at $k=5$ and up to $k=100$.

\subsubsection{Joint Training with the Source Language}

\begin{figure}[tb]
    \centering
    \includegraphics[width=1.0\columnwidth]{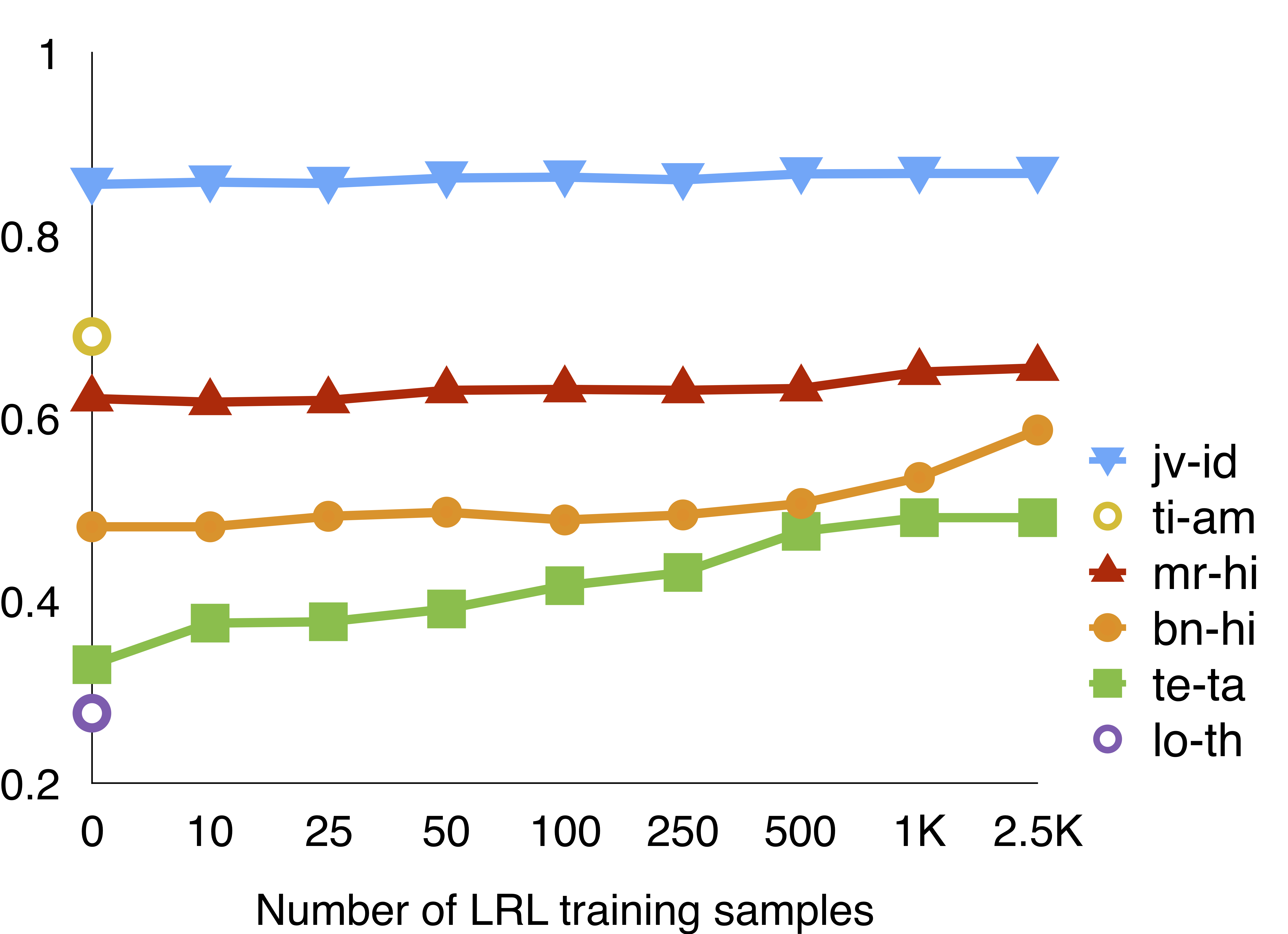}
    \caption{Entity linking accuracy on joint training with different amounts of LRL data (x-axis is not to scale).}
    \label{fig:semisup}
\end{figure}

Although our main focus is an entity linking method that uses zero resources in the source language, we also analyze how the performance of our model improves when jointly trained with a small amount of data in the source LRL. Figure~\ref{fig:semisup} shows the variation of EL accuracy with different amounts, ranging from 10 to 2,500, of LRL samples added to the HRL (\textsc{Manual} setting) data for training the entity similarity encoder. For language pairs that are very closely-related and share the same scripts, such as \textit{jv}-\textit{id} and \textit{mr}-\textit{hi}, there is only a small increase in accuracy even with thousands of source language training samples. In contrast, there is considerable performance improvement for \textit{bn}-\textit{hi} and \textit{te}-\textit{ta}, both of which use phonological transfer since the writing systems are different. We also note that several of our test languages (\textit{ti}, \textit{lo}, \textit{ug}) are so poorly resourced, even on Wikipedia, that there was not enough data for this joint training experiment (shown in Figure~\ref{fig:semisup}). 

\section{Related Work}
\label{sec: related}

We briefly discuss previous work related to various facets of our proposed method.

{\parindent0pt
\textbf{Cross-lingual EL}.\quad The TAC-KBP shared task on entity linking has featured Chinese/Spanish to English EL since 2011~\cite{Ji2011OverviewOT,Ji2014OverviewOT,Ji2017OverviewOT}. Around the same time,
\citeauthor{mcnamee-EtAl:2011:IJCNLP-2011}~\shortcite{mcnamee-EtAl:2011:IJCNLP-2011} introduced cross-lingual EL as a new task and designed a candidate retrieval technique based on Wikipedia language-links. More recently,~\citeauthor{tsai-roth:2016:N16-1}~\shortcite{tsai-roth:2016:N16-1} used word embeddings for EL in 12 languages,~\citeauthor{pan-EtAl:2017:Long2}~\shortcite{pan-EtAl:2017:Long2} use word-for-word translation for in a massive multilingual effort for EL in 282 language pairs and~\citeauthor{TsaiRo18}~\shortcite{TsaiRo18} develop better name translation for improving the performance of existing translation-based EL techniques. The neural model proposed by~\citeauthor{sil2017neural}~\shortcite{sil2017neural} based on multilingual word embeddings and Wikipedia links achieves state-of-the-art results on the TAC2015 dataset.
\smallskip

\textbf{Cross-lingual transfer learning}.\quad \citet{I08-3008} perform transfer to adapt parsers to low-resource languages and~\citet{hwa2005bootstrapping} project parsers from English to other languages that have no syntactic annotation, using parallel texts. 
Neural models for transfer from high-resource to low-resource are used for several tasks including POS tagging, machine translation, and morphological analysis~\cite{zoph-EtAl:2016:EMNLP2016,fang-cohn:2017:Short,cotterell-heigold:2017:EMNLP2017}. Unlike our work, these methods jointly train with a small amount of data in the low-resource language.
\smallskip

\textbf{Phonological representation.}\quad IPA representations of language have been used to identify borrowed words across unrelated languages~\cite{tsvetkov-ammar-dyer:2015:NAACL-HLT,Tsvetkov2016CrossLingualBW}. Perhaps most similar to our work is~\citeauthor{bharadwaj-EtAl:2016:EMNLP2016}~\shortcite{bharadwaj-EtAl:2016:EMNLP2016}, which uses IPA in transferring learned models for named entity recognition to low-resource languages. 

}

\section{Conclusion}

We present \textsc{Pbel}, a method for cross-lingual entity linking that uses zero parallel resources in the language of the input mention. With extensive experiments on nine test languages, we demonstrate its potential for low-resource entity linking across several language families. Our model uses zero-shot transfer from closely-related high-resource languages and improves accuracy by 17\% on average over baseline systems. We also show its ability to transfer across orthographies through phonological representations. 

An immediate future focus for our work could be training a model that predicts the `best' pivot language for a given named entity mention, which can replace the language-specific phylogenetic-distance-based weights used in this work. Further, we currently train individual encoders for each language. Universal multilingual encoders have had success in tasks like translation~\cite{DBLP:journals/corr/JohnsonSLKWCTVW16,DBLP:journals/corr/HaNW16} and can potentially ease the scaling up of our model to a large number of languages.

\section{Acknowledgements}
This work is sponsored by Defense Advanced Research Projects Agency Information Innovation Office (I2O), Program: Low Resource Languages for Emergent Incidents (LORELEI), issued by DARPA/I2O under Contract No. HR0011-15-C-0114. Shruti Rijhwani is supported by a Bloomberg Data Science Ph.D. Fellowship. The authors would also like to thank Abhilasha Ravichander, Aditi Chaudhary, Xinyi Wang, Danish, Deepak Gopinath, Gayatri Bhat, Maria Ryskina, Paul Michel, Shivani Poddar, and Siddharth Dalmia for their reviews while drafting this paper.

\bibliography{acl2018}
\bibliographystyle{aaai}

\end{document}